\documentclass{bmvc2k}
\usepackage{amsmath,amssymb}
\usepackage{bm}
\usepackage{xcolor}
\usepackage{colortbl}
\usepackage{booktabs}
\usepackage{wrapfig}
\usepackage{multirow}
\usepackage{url}

\title{Flatness-aware Curriculum Learning via Adversarial Difficulty}

\addauthor{Hiroaki Aizawa}{hiroaki-aizawa@hiroshima-u.ac.jp}{1}
\addauthor{Yoshikazu Hayashi}{hayashi.yoshikazu.a8@f.gifu-u.ac.jp}{2}

\addinstitution{
Graduate School of Advanced Science and Engineering\\
Hiroshima University\\
Hiroshima, Japan
}
\addinstitution{
Intelligent Production Technology Research \& Development Center for Aerospace\\
Gifu University\\
Gifu, Japan
}

\runninghead{H. Aizawa et al.}{Flattness-aware Curriculum Learning}

\def\acccifar{
\begin{table*}[t]
\small
\setlength{\tabcolsep}{3pt}
\centering
\renewcommand{\arraystretch}{1.1}
\caption{Image classification accuracies (\%) on \texttt{CIFAR10} and \texttt{CIFAR100} (mean $\pm$ std over 3 seeds). \textbf{Curr.} = Curriculum strategy, \textbf{Diff.} = Difficulty measure. Gray rows indicate results using the proposed ADM.}
\begin{tabular}{cc|ccc|ccc}
\toprule[0.4mm]
\multirow{2}{*}{\textbf{Curr.} $v$} & \multirow{2}{*}{\textbf{Diff.} $\xi$}
& \multicolumn{3}{c|}{\texttt{CIFAR10}} 
& \multicolumn{3}{c}{\texttt{CIFAR100}} \\
\cmidrule(lr){3-5} \cmidrule(lr){6-8}
& & ResNet18 & WRN28\_2 & WRN28\_10 & ResNet18 & WRN28\_2 & WRN28\_10 \\
\midrule
\multicolumn{8}{l}{\textbf{Opt.}: SGD} \\
ERM & - & $95.32_{\pm 0.17}$ & $94.86_{\pm 0.09}$ & $96.10_{\pm 0.02}$ & $77.97_{\pm 0.17}$ & $74.87_{\pm 0.03}$ & $\mathbf{81.23}_{\pm 0.05}$ \\
\multirow[c]{2}{*}{SPL} & Loss & $95.33_{\pm 0.19}$ & $94.93_{\pm 0.11}$ & $\mathbf{96.17}_{\pm 0.14}$ & $78.00_{\pm 0.21}$ & $74.60_{\pm 0.34}$ & $80.91_{\pm 0.05}$ \\
& \cellcolor{black!7}ADM & \cellcolor{black!7}$95.33_{\pm 0.12}$ & \cellcolor{black!7}$\mathbf{94.99}_{\pm 0.00}$ & \cellcolor{black!7}$96.12_{\pm 0.02}$ & \cellcolor{black!7}$77.83_{\pm 0.14}$ & \cellcolor{black!7}$\mathbf{75.29}_{\pm 0.31}$ & \cellcolor{black!7}$81.08_{\pm 0.37}$ \\
\multirow[c]{3}{*}{SSPL} & Loss & $95.37_{\pm 0.10}$ & $94.52_{\pm 0.10}$ & $96.01_{\pm 0.10}$ & $78.15_{\pm 0.37}$ & $74.59_{\pm 0.17}$ & $80.92_{\pm 0.26}$ \\
& GDM & $95.23_{\pm 0.10}$ & $94.74_{\pm 0.06}$ & $95.95_{\pm 0.04}$ & $77.92_{\pm 0.02}$ & $74.53_{\pm 0.20}$ & $80.66_{\pm 0.15}$ \\
& \cellcolor{black!7}ADM & \cellcolor{black!7}$\mathbf{95.38}_{\pm 0.11}$ & \cellcolor{black!7}$94.85_{\pm 0.15}$ & \cellcolor{black!7}$96.05_{\pm 0.13}$ & \cellcolor{black!7}$\mathbf{78.78}_{\pm 0.27}$ & \cellcolor{black!7}$75.02_{\pm 0.26}$ & \cellcolor{black!7}$80.80_{\pm 0.12}$ \\
\midrule
\multicolumn{8}{l}{\textbf{Opt.}: SAM} \\
ERM & - & $95.86_{\pm 0.07}$ & $95.32_{\pm 0.09}$ & $96.73_{\pm 0.12}$ & $78.99_{\pm 0.12}$ & $77.04_{\pm 0.13}$ & $82.97_{\pm 0.11}$ \\
\multirow[c]{2}{*}{SPL} & Loss & $96.02_{\pm 0.07}$ & $95.50_{\pm 0.12}$ & $\mathbf{96.84}_{\pm 0.03}$ & $79.04_{\pm 0.21}$ & $76.48_{\pm 0.19}$ & $83.05_{\pm 0.11}$ \\
& \cellcolor{black!7}ADM & \cellcolor{black!7}$96.04_{\pm 0.06}$ & \cellcolor{black!7}$95.41_{\pm 0.11}$ & \cellcolor{black!7}$96.79_{\pm 0.07}$ & \cellcolor{black!7}$78.89_{\pm 0.13}$ & \cellcolor{black!7}$76.79_{\pm 0.19}$ & \cellcolor{black!7}$83.12_{\pm 0.14}$ \\
\multirow[c]{3}{*}{SSPL} & Loss & $95.98_{\pm 0.07}$ & $95.44_{\pm 0.08}$ & $96.78_{\pm 0.04}$ & $79.13_{\pm 0.16}$ & $76.65_{\pm 0.07}$ & $83.17_{\pm 0.08}$ \\
& GDM & $95.89_{\pm 0.06}$ & $95.29_{\pm 0.06}$ & $96.79_{\pm 0.15}$ & $78.85_{\pm 0.06}$ & $76.04_{\pm 0.14}$ & $82.86_{\pm 0.07}$ \\
& \cellcolor{black!7}ADM & \cellcolor{black!7}$\mathbf{96.09}_{\pm 0.04}$ & \cellcolor{black!7}$\mathbf{95.65}_{\pm 0.18}$ & \cellcolor{black!7}$\mathbf{96.84}_{\pm 0.07}$ & \cellcolor{black!7}$\mathbf{79.89}_{\pm 0.27}$ & \cellcolor{black!7}$\mathbf{77.05}_{\pm 0.15}$ & \cellcolor{black!7}$\mathbf{83.24}_{\pm 0.04}$ \\
\bottomrule[0.4mm]
\end{tabular}
\label{tb:acccifar}
\end{table*}
}

\def\finegrainedacc{
\begin{table*}[t]
\small
\setlength{\tabcolsep}{3pt}
\centering
\renewcommand{\arraystretch}{1.1}
\caption{Fine-grained image classification accuracies (\%)}
\begin{tabular}{cc|cccc}
\toprule[0.4mm]
\textbf{Curr.} $v$ & \textbf{Diff.} $\xi$ & \texttt{Cars} & \texttt{Aircraft} & \texttt{CUB} & \texttt{Food} \\
\midrule
\multicolumn{6}{l}{\textbf{Opt.}: SGD} \\
ERM & - & $82.82_{\pm 0.24}$ & $77.97_{\pm 0.57}$ & $61.64_{\pm 0.20}$ & $79.60_{\pm 0.01}$ \\
\multirow[c]{2}{*}{SPL} & Loss & $82.64_{\pm 0.06}$ & $77.55_{\pm 0.43}$ & $61.96_{\pm 0.22}$ & $79.45_{\pm 0.10}$ \\
& \cellcolor{black!7}ADM & \cellcolor{black!7}$82.14_{\pm 0.19}$ & \cellcolor{black!7}$78.44_{\pm 0.54}$ & \cellcolor{black!7}$61.95_{\pm 0.29}$ & \cellcolor{black!7}$79.66_{\pm 0.07}$ \\
\multirow[c]{3}{*}{SSPL} & Loss & $82.88_{\pm 0.27}$ & $77.95_{\pm 0.19}$ & $63.01_{\pm 0.13}$ & $79.38_{\pm 0.14}$ \\
& GDM & $82.64_{\pm 0.19}$ & $77.87_{\pm 0.48}$ & $63.03_{\pm 0.66}$ & $79.61_{\pm 0.19}$ \\
& \cellcolor{black!7}ADM & \cellcolor{black!7}$\textbf{83.89}_{\pm 0.09}$ & \cellcolor{black!7}$\textbf{78.80}_{\pm 0.35}$ & \cellcolor{black!7}$\textbf{63.63}_{\pm 0.11}$ & \cellcolor{black!7}$\textbf{80.20}_{\pm 0.08}$ \\
\midrule
\multicolumn{6}{l}{\textbf{Opt.}: SAM} \\
ERM & - & $83.24_{\pm 0.32}$ & $78.58_{\pm 0.40}$ & $62.02_{\pm 0.40}$ & $80.91_{\pm 0.13}$ \\
\multirow[c]{2}{*}{SPL} & Loss & $82.97_{\pm 0.55}$ & $78.73_{\pm 0.61}$ & $62.10_{\pm 0.73}$ & $80.89_{\pm 0.05}$ \\
& \cellcolor{black!7}ADM & \cellcolor{black!7}$83.26_{\pm 0.50}$ & \cellcolor{black!7}$78.44_{\pm 0.12}$ & \cellcolor{black!7}$62.65_{\pm 0.22}$ & \cellcolor{black!7}$80.93_{\pm 0.11}$ \\
\multirow[c]{3}{*}{SSPL} & Loss & $83.64_{\pm 0.56}$ & $78.72_{\pm 0.42}$ & $63.39_{\pm 0.56}$ & $81.16_{\pm 0.11}$ \\
& GDM & $83.33_{\pm 0.19}$ & $78.36_{\pm 0.04}$ & $63.28_{\pm 0.27}$ & $80.95_{\pm 0.09}$ \\
& \cellcolor{black!7}ADM & \cellcolor{black!7}$\textbf{83.85}_{\pm 0.05}$ & \cellcolor{black!7}$\textbf{79.32}_{\pm 0.41}$ & \cellcolor{black!7}$\textbf{63.44}_{\pm 0.24}$ & \cellcolor{black!7}$\textbf{82.01}_{\pm 0.09}$ \\
\bottomrule[0.4mm]
\end{tabular}
\label{tb:fine_grained_acc}
\end{table*}
}

\def\dgacc{
\begin{table*}[t]
\small
\setlength{\tabcolsep}{3pt}
\centering
\renewcommand{\arraystretch}{1.1}
\caption{Domain generalization (DG) accuracies (\%)}
\begin{tabular}{cc|ccccc}
\toprule[0.4mm]
\textbf{Curr.} $v$ & \textbf{Diff.} $\xi$ & \texttt{PACS} & \texttt{VLCS} & \texttt{OH} & \texttt{TI} & Avg. \\
\midrule
\multicolumn{7}{l}{\textbf{Opt.}: Adam} \\
ERM & - & $\textbf{82.16}_{\pm 0.77}$ & $\textbf{75.35}_{\pm 1.02}$ & $55.32_{\pm 0.70}$ & $42.07_{\pm 0.35}$ & $63.72$ \\
\multirow[c]{2}{*}{SPL} & Loss & $80.36_{\pm 1.37}$ & $74.95_{\pm 1.48}$ & $55.18_{\pm 0.57}$ & $39.64_{\pm 0.48}$ & $62.53$ \\
& \cellcolor{black!7}ADM & \cellcolor{black!7}$80.82_{\pm 0.04}$ & \cellcolor{black!7}$75.08_{\pm 0.11}$ & \cellcolor{black!7}$54.27_{\pm 0.32}$ & \cellcolor{black!7}$40.05_{\pm 1.90}$ & \cellcolor{black!7}$62.55$ \\
\multirow[c]{3}{*}{SSPL} & Loss & $80.08_{\pm 0.78}$ & $73.86_{\pm 0.48}$ & $46.43_{\pm 0.95}$ & $38.92_{\pm 1.41}$ & $59.82$ \\
& GDM & $79.90_{\pm 0.41}$ & $73.78_{\pm 0.37}$ & $56.01_{\pm 0.60}$ & $40.59_{\pm 1.68}$ & $62.57$ \\
& \cellcolor{black!7}ADM & \cellcolor{black!7}$81.11_{\pm 0.69}$ & \cellcolor{black!7}$75.03_{\pm 1.03}$ & \cellcolor{black!7}$\textbf{57.10}_{\pm 0.37}$ & \cellcolor{black!7}$\textbf{42.81}_{\pm 0.46}$ & \cellcolor{black!7}$\textbf{64.01}$ \\
\midrule
\multicolumn{7}{l}{\textbf{Opt.}: SAM} \\
ERM & - & $84.66_{\pm 0.38}$ & $76.84_{\pm 0.39}$ & $60.76_{\pm 0.17}$ & $44.55_{\pm 0.50}$ & $66.70$ \\
\multirow[c]{2}{*}{SPL} & Loss & $84.54_{\pm 0.95}$ & $76.88_{\pm 0.34}$ & $\textbf{61.19}_{\pm 0.31}$ & $45.36_{\pm 0.57}$ & $66.99$ \\
& \cellcolor{black!7}ADM & \cellcolor{black!7}$\textbf{85.43}_{\pm 0.40}$ & \cellcolor{black!7}$\textbf{77.13}_{\pm 0.38}$ & \cellcolor{black!7}$59.32_{\pm 0.16}$ & \cellcolor{black!7}$43.19_{\pm 2.86}$ & \cellcolor{black!7}$66.27$ \\
\multirow[c]{3}{*}{SSPL} & Loss & $83.60_{\pm 0.31}$ & $76.79_{\pm 0.53}$ & $53.33_{\pm 0.87}$ & $42.63_{\pm 0.63}$ & $64.09$ \\
& GDM & $82.47_{\pm 0.67}$ & $75.06_{\pm 0.32}$ & $58.36_{\pm 0.18}$ & $40.39_{\pm 2.73}$ & $64.07$ \\
& \cellcolor{black!7}ADM & \cellcolor{black!7}$83.88_{\pm 1.32}$ & \cellcolor{black!7}$76.87_{\pm 0.36}$ & \cellcolor{black!7}$60.88_{\pm 0.26}$ & \cellcolor{black!7}$\textbf{46.92}_{\pm 0.77}$ & \cellcolor{black!7}$\textbf{67.14}$ \\
\midrule
\multicolumn{7}{l}{\textbf{Flatness-aware DG approach}: SAGM~\cite{sagm}} \\
ERM & - & $83.72_{\pm 0.74}$ & $76.31_{\pm 0.47}$ & $60.34_{\pm 0.21}$ & $46.00_{\pm 1.61}$ & $66.59$ \\
\bottomrule[0.4mm]
\end{tabular}
\label{tb:dg}
\end{table*}
}

\def\acchessian{
\begin{wraptable}{r}{0.5\linewidth}
\small
\renewcommand{\arraystretch}{1.1}
\centering
\caption{Maximum Hessian eigenvalue. Lower value denotes flat minima.}
\begin{tabular}{lcccc}
\toprule[0.4mm]
\multirow{2}{*}{\textbf{Curr.} $v$} & \multicolumn{2}{c}{\texttt{CIFAR10}} & \multicolumn{2}{c}{\texttt{CIFAR100}} \\
\cmidrule(lr){2-3} \cmidrule(lr){4-5}
& SGD & SAM & SGD & SAM \\
\midrule
ERM & $8.25$ & $3.03$ & $\textbf{22.87}$ & $9.47$ \\
SPL & $\textbf{6.17}$ & $\textbf{2.35}$ & $33.97$ & $\textbf{7.49}$ \\
SSPL & $8.55$ & $6.11$ & $78.78$ & $19.92$ \\
\bottomrule[0.4mm]
\end{tabular}
\label{tb:acc_hessian}
\end{wraptable}
}

\def\cifarc{
\begin{table*}[t]
\small
\setlength{\tabcolsep}{3pt}
\centering
\renewcommand{\arraystretch}{1.1}
\caption{Image classification accuracies (\%) on \texttt{CIFAR10-C} and \texttt{CIFAR100-C}.}
\begin{tabular}{cc|ccc|ccc}
\toprule[0.4mm]
\multirow{2}{*}{\textbf{Curr.} $v$} & \multirow{2}{*}{\textbf{Diff.} $\xi$} 
& \multicolumn{3}{c|}{\texttt{CIFAR10-C}} 
& \multicolumn{3}{c}{\texttt{CIFAR100-C}} \\
\cmidrule(lr){3-5} \cmidrule(lr){6-8}
& & ResNet18 & WRN28\_2 & WRN28\_10 & ResNet18 & WRN28\_2 & WRN28\_10 \\
\midrule
\multicolumn{8}{l}{\textbf{Opt.}: SGD} \\
ERM & - & $74.08_{\pm 0.38}$ & $\textbf{73.61}_{\pm 0.57}$ & $76.37_{\pm 0.09}$ & $48.50_{\pm 0.08}$ & $45.05_{\pm 0.12}$ & $\textbf{51.78}_{\pm 0.13}$ \\
\multirow[c]{2}{*}{SPL} & Loss & $74.14_{\pm 0.10}$ & $73.35_{\pm 0.07}$ & $\textbf{76.56}_{\pm 0.28}$ & $48.64_{\pm 0.11}$ & $\textbf{45.74}_{\pm 0.41}$ & $\textbf{51.78}_{\pm 0.14}$ \\
& \cellcolor{black!7}ADM & \cellcolor{black!7}$74.24_{\pm 0.17}$ & \cellcolor{black!7}$73.19_{\pm 0.36}$ & \cellcolor{black!7}$76.02_{\pm 0.45}$ & \cellcolor{black!7}$48.78_{\pm 0.16}$ & \cellcolor{black!7}$45.61_{\pm 0.41}$ & \cellcolor{black!7}$51.44_{\pm 0.26}$ \\
\multirow[c]{3}{*}{SSPL} & Loss & $74.03_{\pm 0.25}$ & $72.93_{\pm 0.15}$ & $75.42_{\pm 0.29}$ & $48.66_{\pm 0.04}$ & $45.10_{\pm 0.29}$ & $51.58_{\pm 0.16}$ \\
& GDM & $\textbf{74.26}_{\pm 0.43}$ & $73.22_{\pm 0.43}$ & $75.88_{\pm 0.37}$ & $48.31_{\pm 0.46}$ & $45.04_{\pm 0.02}$ & $51.55_{\pm 0.48}$ \\
& \cellcolor{black!7}ADM & \cellcolor{black!7}$74.15_{\pm 0.21}$ & \cellcolor{black!7}$73.39_{\pm 0.40}$ & \cellcolor{black!7}$75.43_{\pm 0.29}$ & \cellcolor{black!7}$\textbf{49.03}_{\pm 0.12}$ & \cellcolor{black!7}$45.34_{\pm 0.23}$ & \cellcolor{black!7}$51.43_{\pm 0.19}$ \\
\midrule
\multicolumn{8}{l}{\textbf{Opt.}: SAM} \\
ERM & - & $75.78_{\pm 0.39}$ & $74.60_{\pm 0.28}$ & $77.58_{\pm 0.29}$ & $49.91_{\pm 0.14}$ & $46.76_{\pm 0.32}$ & $53.92_{\pm 0.12}$ \\
\multirow[c]{2}{*}{SPL} & Loss & $75.94_{\pm 0.16}$ & $74.38_{\pm 0.55}$ & $\textbf{77.90}_{\pm 0.22}$ & $49.99_{\pm 0.28}$ & $46.60_{\pm 0.34}$ & $\textbf{54.45}_{\pm 0.12}$ \\
& \cellcolor{black!7}ADM & \cellcolor{black!7}$75.36_{\pm 0.49}$ & \cellcolor{black!7}$73.65_{\pm 0.28}$ & \cellcolor{black!7}$77.67_{\pm 0.21}$ & \cellcolor{black!7}$50.06_{\pm 0.14}$ & \cellcolor{black!7}$47.34_{\pm 0.29}$ & \cellcolor{black!7}$53.92_{\pm 0.27}$ \\
\multirow[c]{3}{*}{SSPL} & Loss & $75.56_{\pm 0.29}$ & $73.48_{\pm 0.25}$ & $77.46_{\pm 0.16}$ & $50.29_{\pm 0.34}$ & $46.91_{\pm 0.38}$ & $54.14_{\pm 0.12}$ \\
& GDM & $75.21_{\pm 0.49}$ & $73.73_{\pm 0.34}$ & $77.28_{\pm 0.36}$ & $49.74_{\pm 0.22}$ & $46.48_{\pm 0.07}$ & $53.77_{\pm 0.17}$ \\
& \cellcolor{black!7}ADM & \cellcolor{black!7}$\textbf{76.00}_{\pm 0.26}$ & \cellcolor{black!7}$\textbf{74.62}_{\pm 0.29}$ & \cellcolor{black!7}$77.60_{\pm 0.54}$ & \cellcolor{black!7}$\textbf{50.95}_{\pm 0.37}$ & \cellcolor{black!7}$\textbf{47.56}_{\pm 0.37}$ & \cellcolor{black!7}$54.40_{\pm 0.21}$ \\
\bottomrule[0.4mm]
\end{tabular}
\label{tb:cifar_c}
\end{table*}
}
\def\overview{
\begin{figure}[t]
\centering
\includegraphics[width=0.8\linewidth]{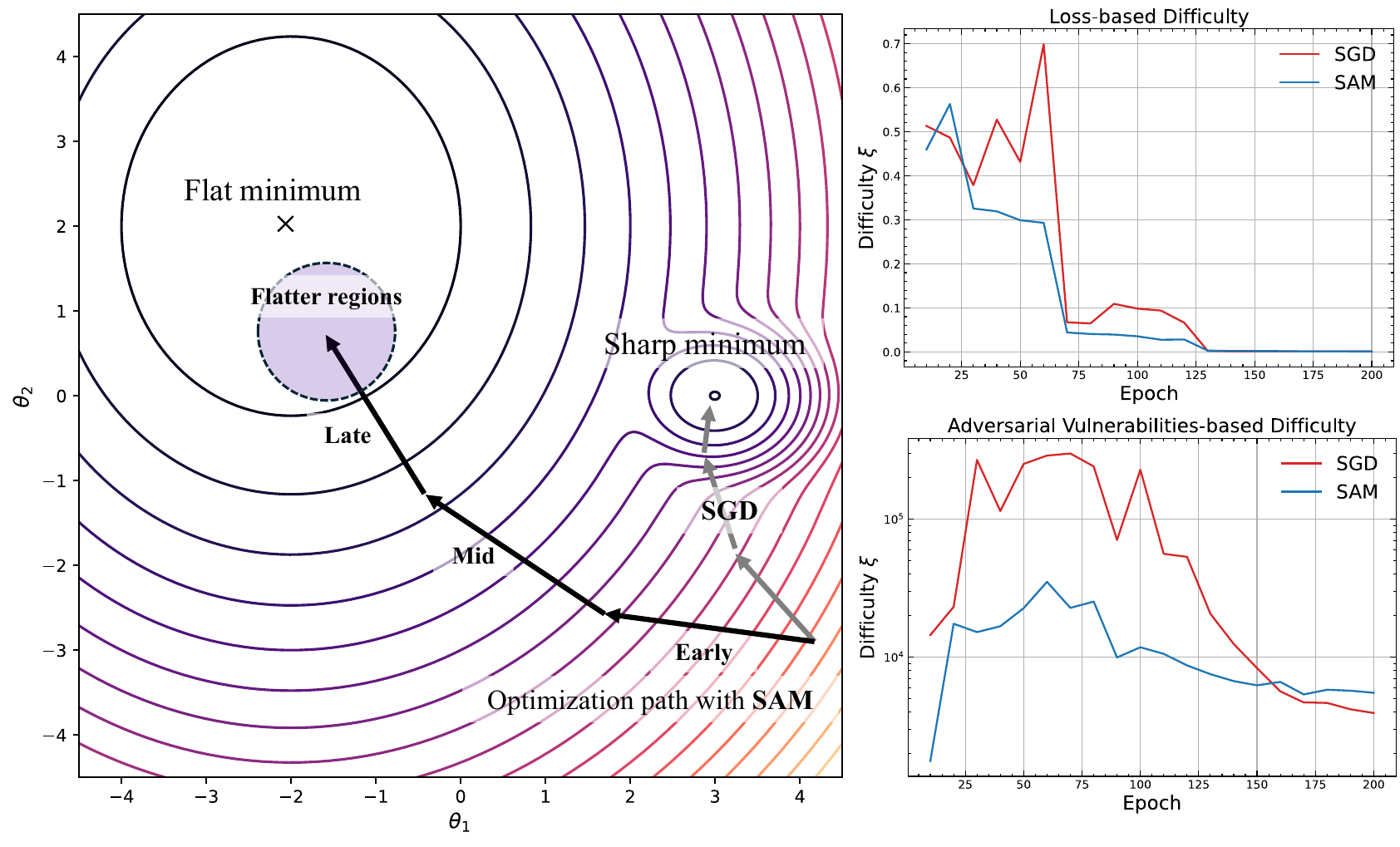}
\caption{Illustration of the optimization trajectory toward a flat minimum (left) and visualizations of the loss-based difficulty (top-right) and our proposed measure (bottom-right). As the optimization trajectory moves toward the flat minimum in the late phase of training (approximately 125-200 epochs), the loss-based difficulty becomes uniformly small, making it hard to evaluate examples for CL. In contrast, our difficulty, measured as the normalized loss gap between clean and adversarial samples, remains informative signal for CL.}
\label{fig:overview}
\end{figure}
}

\def\visdiff{
\begin{figure*}[t]
\centering
\begin{minipage}[b]{0.32\textwidth}
  \centering
  \includegraphics[width=\textwidth]{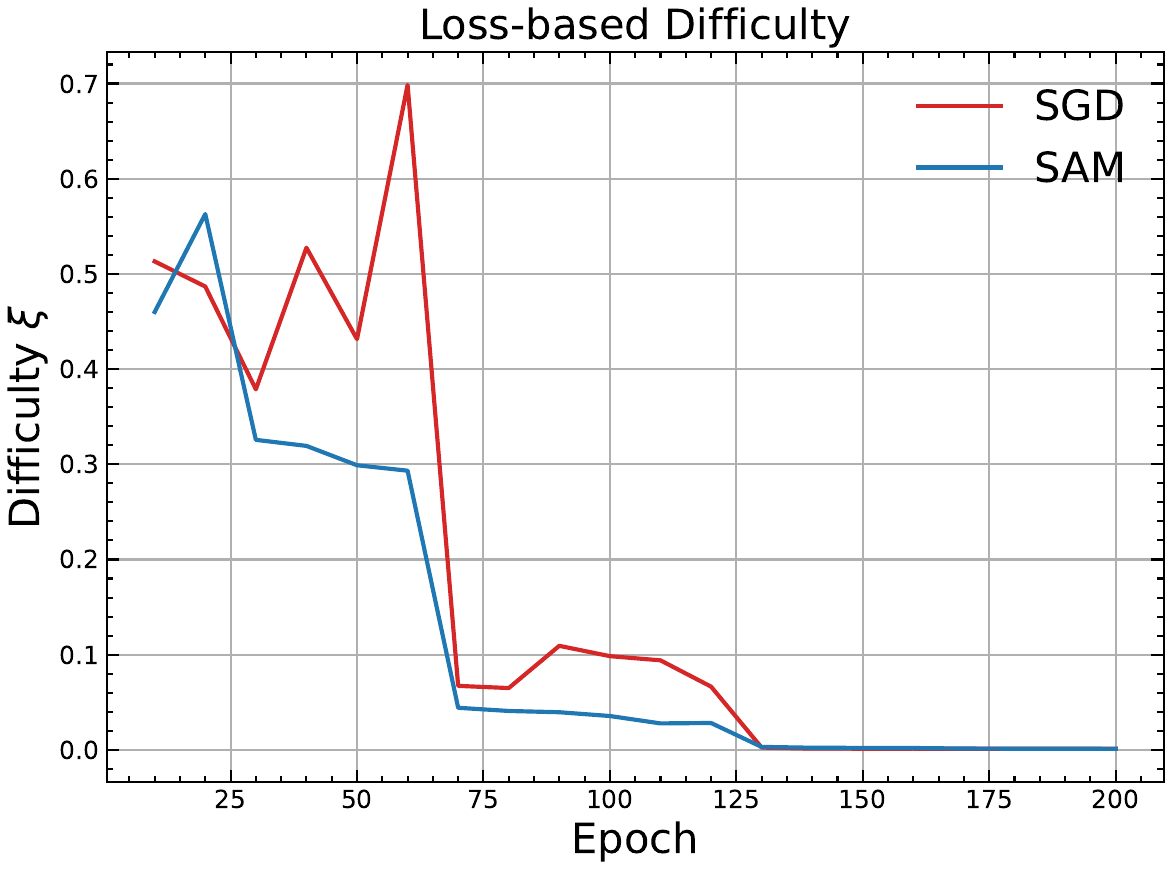}
\end{minipage}
\hfill
\begin{minipage}[b]{0.32\textwidth}
  \centering
  \includegraphics[width=\textwidth]{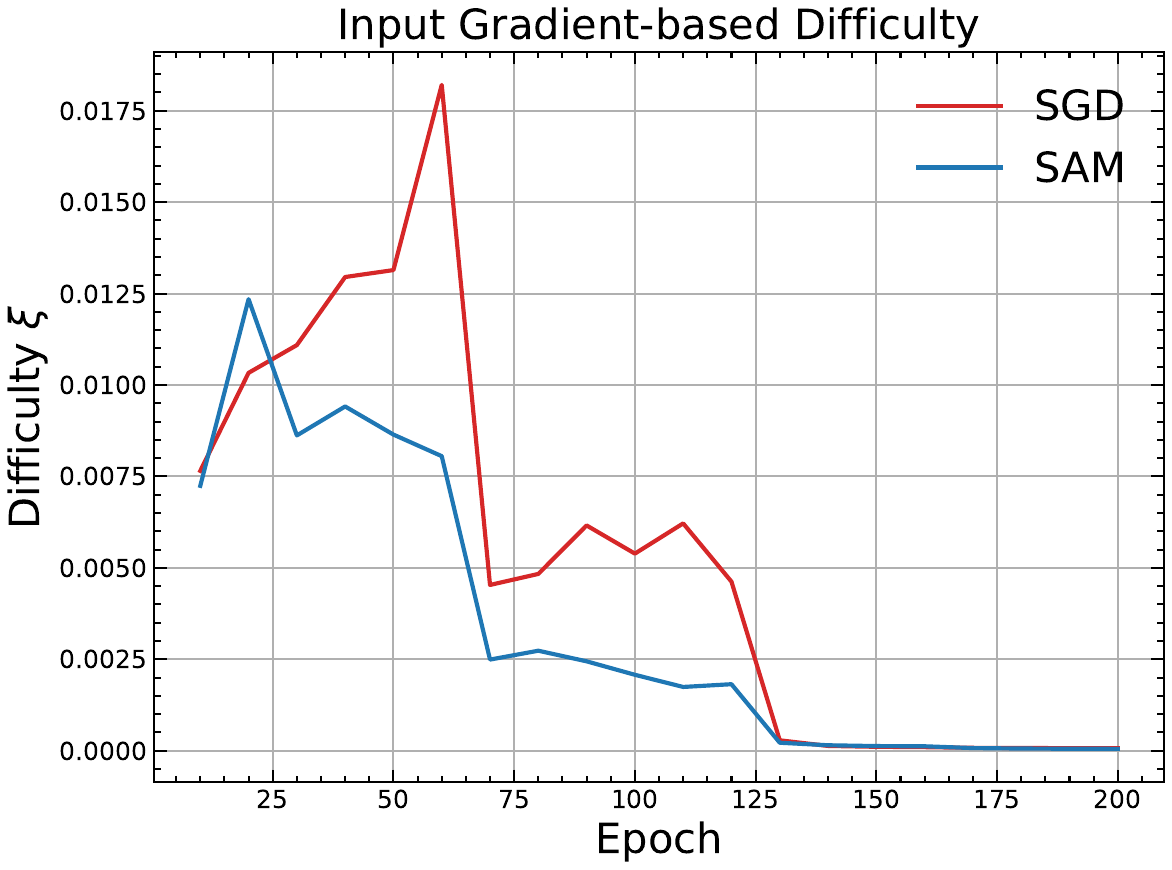}
\end{minipage}
\hfill
\begin{minipage}[b]{0.32\textwidth}
  \centering
  \includegraphics[width=\textwidth]{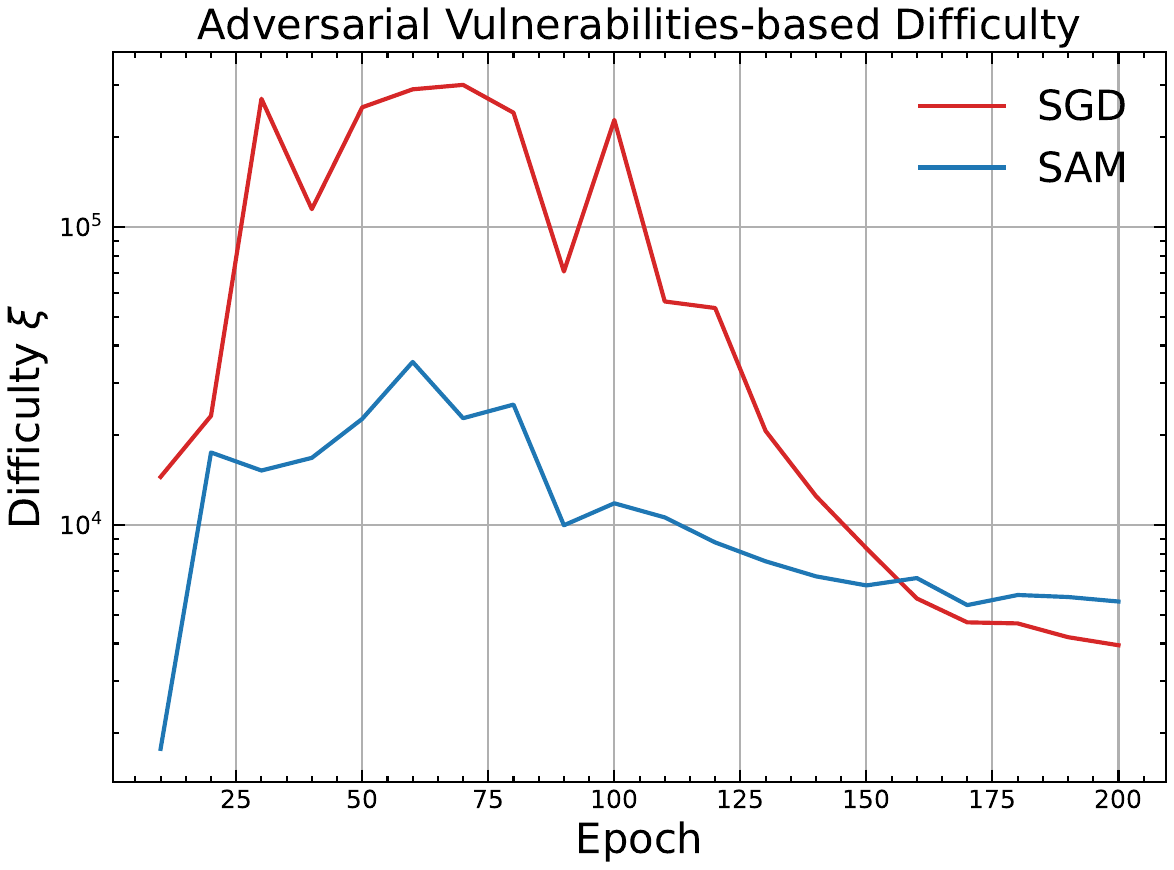}
\end{minipage}
\caption{Visualization of the difficulties for curriculum learning during the training on \texttt{CIFAR10} with SGD and SAM.}
\label{fig:sharp_flat_difficulty}
\end{figure*}
}

\def\losslandscape{
\begin{figure*}[t]
\centering

\begin{minipage}[b]{0.23\textwidth}
  \centering
  \includegraphics[width=\textwidth]{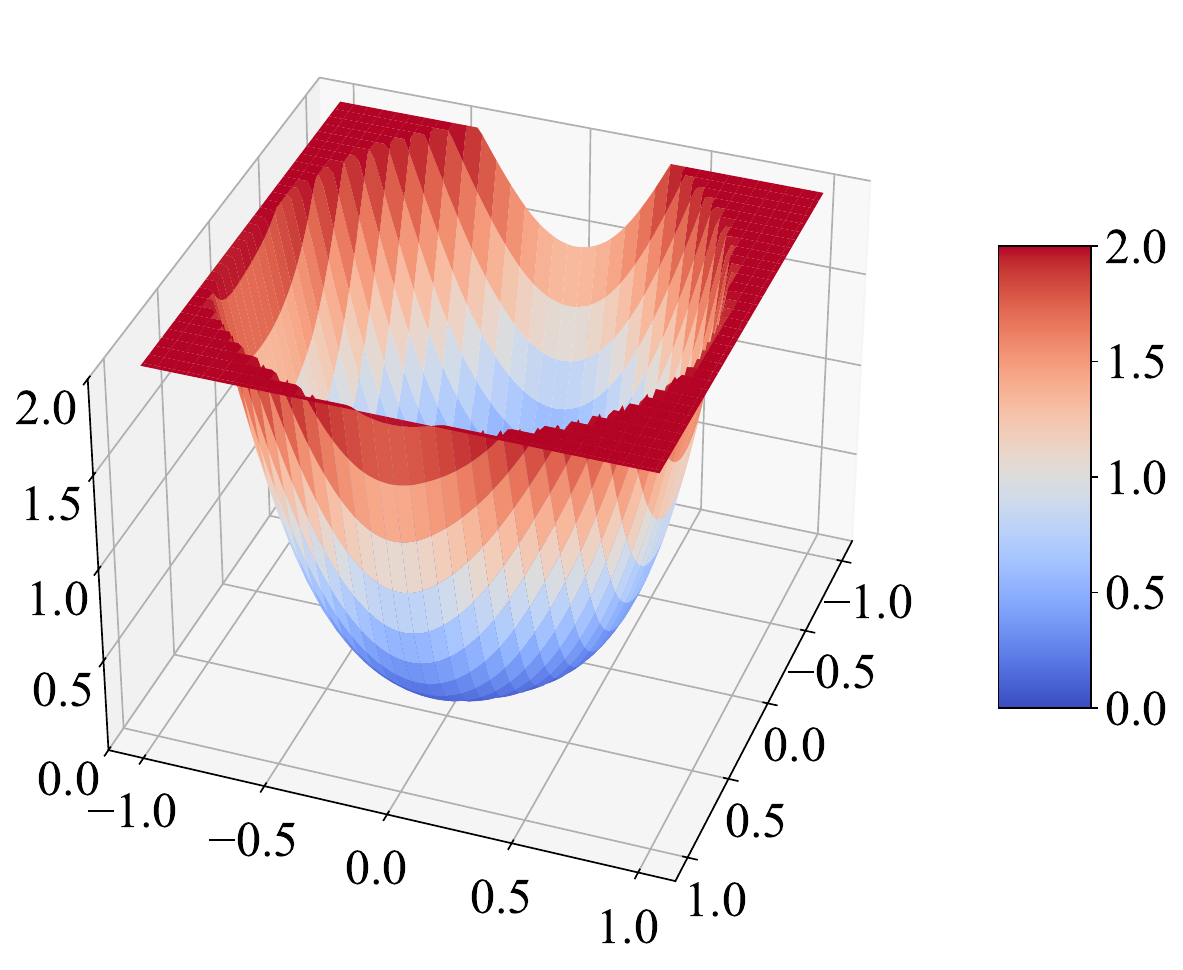}
  
  \vspace{0.3em}
  {\scriptsize ERM w/ SGD}
\end{minipage}
\hfill
\begin{minipage}[b]{0.23\textwidth}
  \centering
  \includegraphics[width=\textwidth]{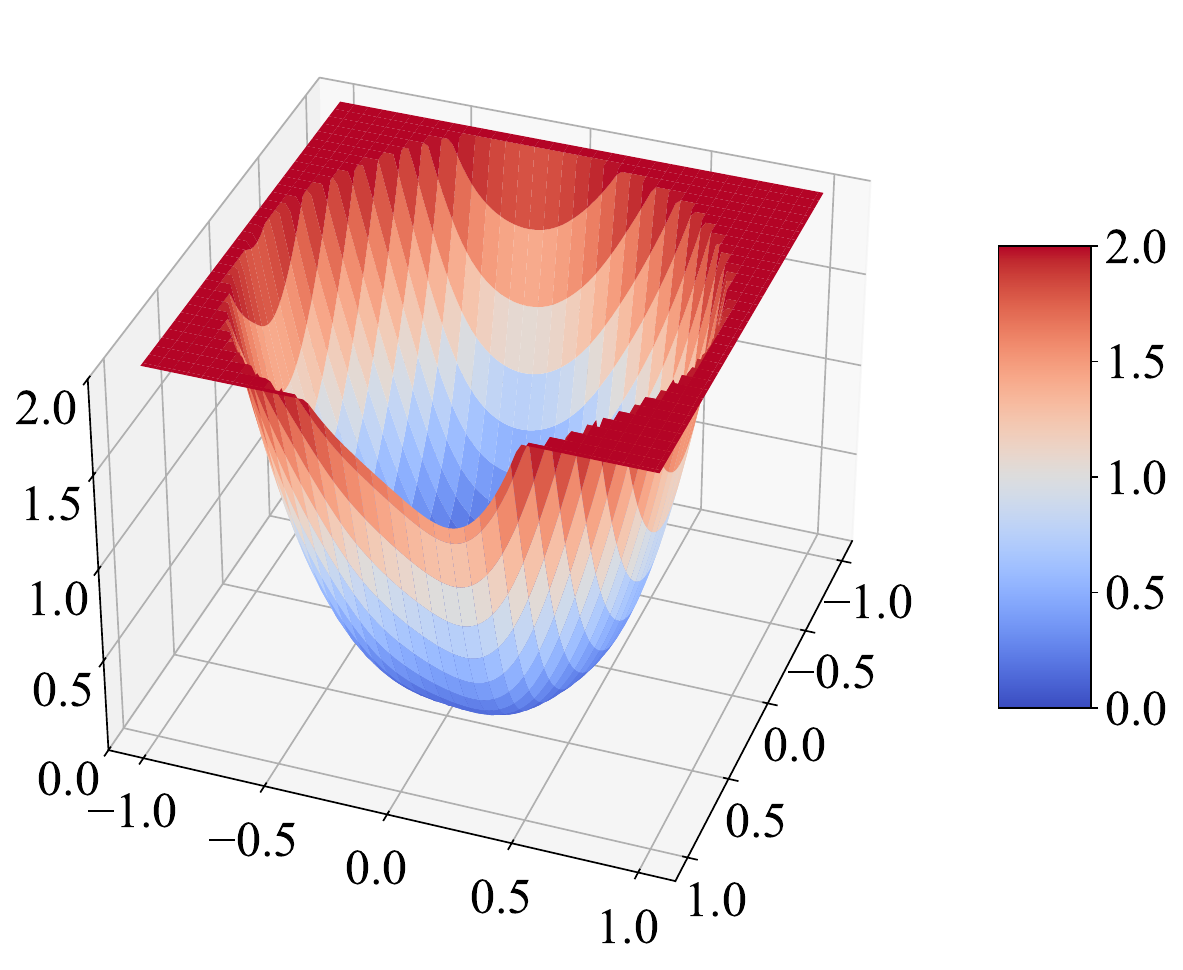}
  
  \vspace{0.3em}
  {\scriptsize ERM w/ SAM}
\end{minipage}
\hfill
\begin{minipage}[b]{0.23\textwidth}
  \centering
  \includegraphics[width=\textwidth]{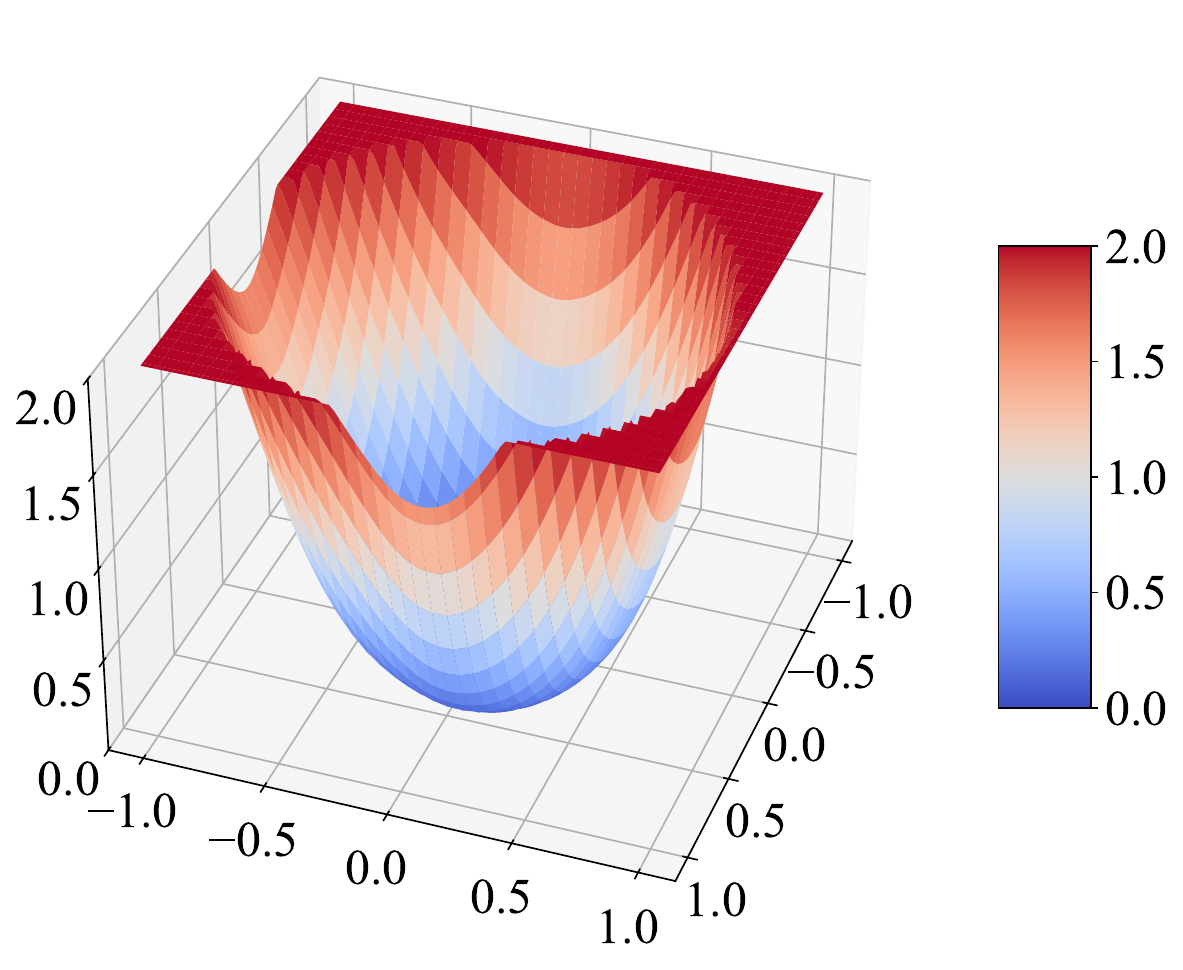}
  
  \vspace{0.3em}
  {\scriptsize ADM-based SPL w/ SAM}
\end{minipage}
\hfill
\begin{minipage}[b]{0.23\textwidth}
  \centering
  \includegraphics[width=\textwidth]{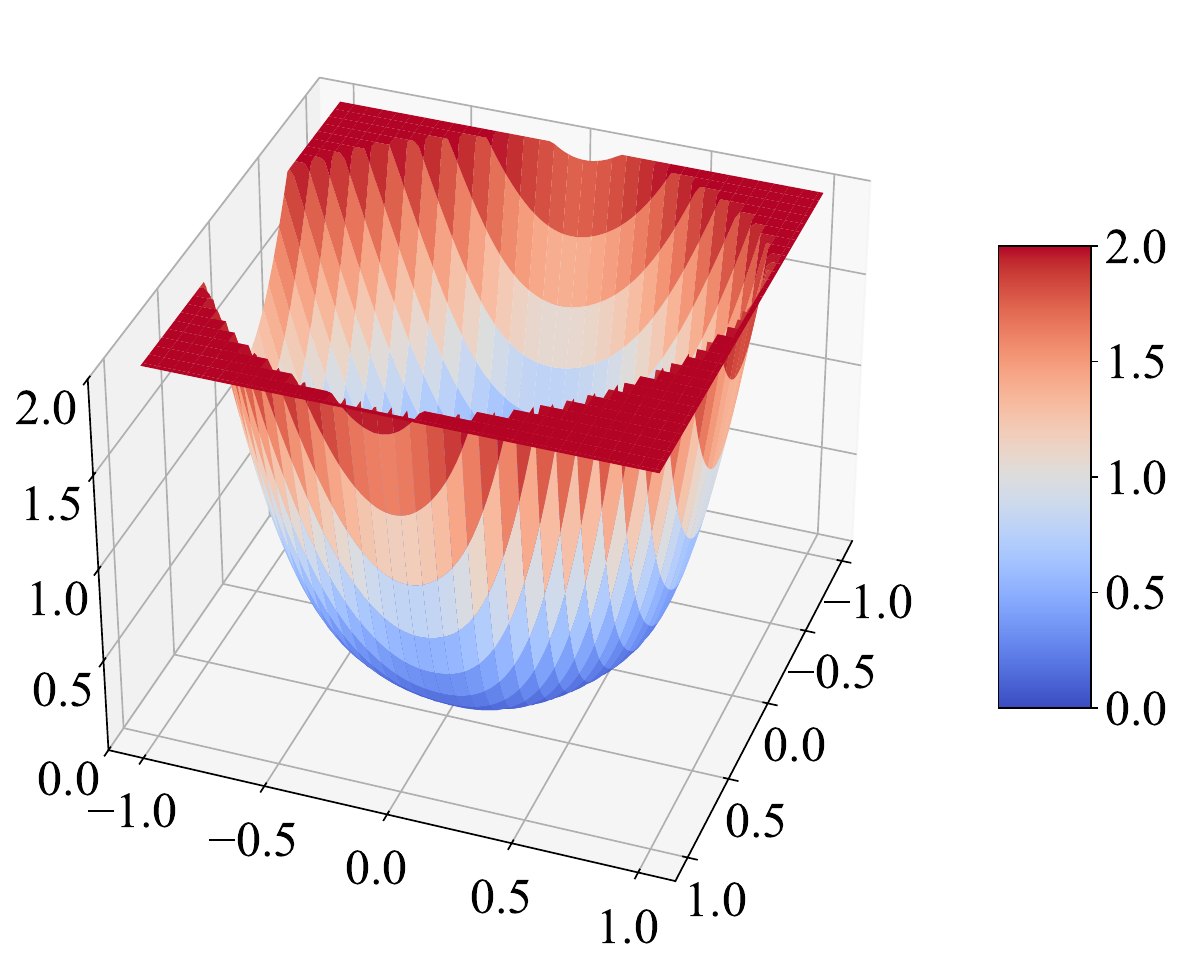}
  
  \vspace{0.3em}
  {\scriptsize ADM-based SSPL w/ SAM}
\end{minipage}
\caption{Visualization of loss landscapes on \texttt{CIFAR-10}.}
\label{fig:loss_landscape}
\end{figure*}
}

\begin{document}

\maketitle

\begin{abstract}
Neural networks trained by empirical risk minimization often suffer from overfitting, especially to specific samples or domains, which leads to poor generalization. Curriculum Learning (CL) addresses this issue by selecting training samples based on the difficulty. From the optimization perspective, methods such as Sharpness-Aware Minimization (SAM) improve robustness and generalization by seeking flat minima. However, combining CL with SAM is not straightforward. In flat regions, both the loss values and the gradient norms tend to become uniformly small, which makes it difficult to evaluate sample difficulty and design an effective curriculum. To overcome this problem, we propose the Adversarial Difficulty Measure (ADM), which quantifies adversarial vulnerability by leveraging the robustness properties of models trained toward flat minima. Unlike loss- or gradient-based measures, which become ineffective as training progresses into flatter regions, ADM remains informative by measuring the normalized loss gap between original and adversarial examples. We incorporate ADM into CL-based training with SAM to dynamically assess sample difficulty. We evaluated our approach on image classification tasks, fine-grained recognition, and domain generalization. The results demonstrate that our method preserves the strengths of both CL and SAM while outperforming existing curriculum-based and flatness-aware training strategies. 
\end{abstract}

\overview

\section{Introduction}
\label{sec:outline}
Neural networks are typically trained based on Empirical Risk Minimization (ERM), which aims to minimize the loss over training data. While effective in many settings, ERM often leads to overfitting. In such cases, the model fits spurious patterns in the training data and fails to generalize well. This limitation has become a fundamental challenge in machine learning and computer vision.

Among various strategies to improve generalization, Curriculum Learning (CL)~\cite{cl,spl,sspl,clsurvey}, has gained attention as an promising method that improves training dynamics by selecting samples in a curriculum determined by their difficulty, typically from simpler to more complex examples. A representative instance of CL is Self-Paced Learning (SPL)~\cite{spl}, which defines sample difficulty based on loss values and prioritizes low-loss examples during the early stages of training. Symmetric SPL (SSPL)~\cite{sspl} further extends this idea by introducing symmetric curriculum scheduler and gradient difficulty measure which utilizes the norm of the input gradient. While these approaches have demonstrated strong performance across various tasks, they remain a limitation in that their difficulty estimates rely either on the loss or its local gradient.

From an optimization perspective, flat minima~\cite{flatminima} are known to contribute to better generalization and robustness~\cite{largebatchflat,sam,asam,samat,samlabelnoise}. Flat minima correspond to regions where the loss remains relatively insensitive to perturbations in the parameters. Models that converge to such regions have been shown both theoretically and empirically to be more resilient to label noise~\cite{sam,samlabelnoise}, adversarial attacks~\cite{samat}, and distribution shifts~\cite{swad,sagm}. To promote convergence to flat minimas, several optimization methods and regularization techniques~\cite{gradnorm} have been proposed. Among them, Sharpness-Aware Minimization (SAM)~\cite{sam} has demonstrated strong performance across various benchmarks.

To further improve generalization, we investigate combining CL with flatness-aware optimization methods like SAM. However, this integration is not straightforward. As training progresses and the model converges toward flatter regions, both the loss values and input gradient norms tend to become uniformly small. In particular, while loss~\cite{spl}- and gradient~\cite{sspl}-based difficulty measures are informative during the early phase of training, their effectiveness rapidly degrades in the intermediate and final phases under flat minima. This trend makes it increasingly difficult to estimate sample difficulty reliably, thereby diminishing the efficacy of curricula based on SPL and SSPL. Figure~\ref{fig:overview} illustrates this phenomenon. Consequently, it becomes difficult to fully leverage the benefits of CL in sample selection and those of SAM in optimization.

To bridge this gap, we introduce a novel difficulty measure that leverages the observation that flat minima tend to exhibit increased robustness against adversarial perturbations~\cite{samat}. Specifically, we propose the Adversarial Difficulty Measure (ADM), which quantifies adversarial vulnerability by evaluating the normalized loss difference between a clean sample and its adversarial counterpart~\cite{ae}. By using ADM to construct the curriculum, we enable reliable sample difficulty estimation even as the model converges toward flatter regions of the loss landscape. We evaluate our method across image classification, fine-grained recognition, and domain generalization scenarios. The results show that our method preserves the strengths of both CL and SAM while outperforming existing curriculum-based and flatness-aware training strategies.

\section{Method}
\label{sec:method}

\subsection{Curriculum-Based Weighted Loss}
\label{sec:weighted_loss}

We consider a $K$-class classification task. Let $\mathcal{D} = \{(\boldsymbol{x}_i, y_i)\}_{i=1}^N$ be a $N$ samples labeled dataset consisting of $d$-dimentional input data $\boldsymbol{x}_i \in \mathbb{R}^d$ and its corresponding label $y_i \in \{1, \ldots, K\}$. We aim to learn a classification model $f(\cdot; \theta)$ with parameters $\theta$. In curriculum-based learning, we assign a weight vector $\boldsymbol{v} \in \mathbb{R}^N$ indicating as the curriculum, and train the model by minimizing the weighted loss:
\begin{align}
\mathcal{L}_{\mathrm{CL}}(\theta, \boldsymbol{v}) = \sum_{i=1}^{N} v_i \ell\bigl(f(\boldsymbol{x}_i;\theta), y_i\bigr),
\label{eq:loss_spl}
\end{align}
where $\ell$ denotes the cross-entropy loss and $v_i$ indicate the $i$-th sample weight. If all weights are set to $v_i = 1$, the objective is Empirical Risk Minimization (ERM).

The weights ${v_i}$ are dynamically updated during training according to the estimated difficulty of each sample, denoted by $\xi_i$. In SPL~\cite{spl}, difficulty is typically estimated based on the cross entropy loss itself, while Symmetric SPL (SSPL)~\cite{sspl} uses the norm of the input gradient as a proxy, as the Gradient Difficulty Measure (DGM). Based on these difficulty measures, the curriculum is updated over time so that training starts with easier samples and gradually incorporates harder ones in later stages.

\subsection{Seeking Flat Minima in Curriculum Learning}
To seek such flat minima from the weighted loss $\mathcal{L}_{\mathrm{CL}}$ defined in Eq.\eqref{eq:loss_spl}, Sharpness-Aware Minimization (SAM)~\cite{sam} formulates the worst-case loss within a neighborhood of the current parameter $\theta$ as:
\begin{align}
\widetilde{\mathcal{L}}_{\mathrm{CL}}(\theta, \boldsymbol{v})=\max _{\|\epsilon\|_2 \leq \rho} \mathcal{L}_{\mathrm{CL}}(\theta+\epsilon, \boldsymbol{v}),
\label{eq:loss_spl_sam}    
\end{align}
where $\epsilon \in \mathbb{R}^{|\theta|}$ is a perturbation in the perameter space and $\rho > 0$ defines the radius of the perturbation ball in Euclidean norm. A larger $\rho$ encourages exploration of wider neighborhoods in the parameter space, promoting flatter region.

Directly optimizing Eq.~\eqref{eq:loss_spl_sam} is computationally infeasible. Therefore, SAM approximates the perturbation $\epsilon^*$ that locally maximizes the loss via a first-order Taylor expansion:
\begin{align}
\epsilon^* \approx \rho \frac{\nabla_\theta \mathcal{L}_{\mathrm{CL}}(\theta, \boldsymbol{v})}{\left\|\nabla_\theta \mathcal{L}_{\mathrm{CL}}(\theta, \boldsymbol{v})\right\|_2}.
\end{align}

The model parameters are then updated using the gradient evaluated at the perturbed parameter:
\begin{align}
\theta \leftarrow \theta-\eta \nabla_\theta \mathcal{L}_{\mathrm{CL}}\left(\theta+\epsilon^{*}, \boldsymbol{v}\right),
\label{eq:update_sam}
\end{align}
where $\eta$ is the learning rate. This two-step propagation allows the model to optimize toward flat minima while maintaining the sample selection mechanism defined by the weighted loss in Eq.~\eqref{eq:loss_spl}.

\subsection{Adversarial Difficulty Measure}
The SAM update rule defined in Eq.\eqref{eq:update_sam} enables optimization toward flatter minima. In such regions, the loss function becomes less sensitive to small parameter perturbations, which is associated with reduced gradient norms, lower Hessian eigenvalues, and improved robustness to input perturbations such as adversarial noise. While these properties are beneficial for generalization, they also make difficulty estimation based on loss values (as in SPL) or input gradients (as in SSPL) unreliable near flatter region. As the loss surface becomes locally flat, both loss values and gradient norms tend to diminish, which undermines their effectiveness as difficulty indicators.

To address this issue, we introduce the Adversarial Difficulty Measure (ADM), which estimates sample difficulty based on adversarial vulnerability. Our idea is motivated from the observation that flatter minima generally confer greater robustness to adversarial perturbations~\cite{samat}. Specifically, we use the gradient of the loss with respect to the input, which is already computed during the SAM backward pass, to generate adversarial examples via a single-step Fast Gradient Sign Method (FGSM)~\cite{ae}:
\begin{align}
\boldsymbol{x}_i^{\text{adv}} = 
\boldsymbol{x}_i + \alpha \cdot
\mathrm{sign}\Bigl(
\nabla_{\boldsymbol{x}_i} \ell\bigl(f(\boldsymbol{x}_i;\theta), y_i\bigr)
\Bigr),
\end{align}
where $\alpha$ is a scalar that controls the strength of the adversarial noise, and $\mathrm{sign}(\cdot)$ denotes the element-wise sign function. In this paper, we use fixed perturbation strength of $\alpha = 8/255$.

Rather than using adversarial examples solely for robustness, we exploit the difference in loss between the clean and adversarial samples as a proxy for difficulty. The difficulty $\xi_{\mathrm{adv}}(\boldsymbol{x}_i)$ is defined as:
\begin{align}
\xi_{\mathrm{adv}}\left(\boldsymbol{x}_i\right)=\frac{| \ell\bigl(f\left(\boldsymbol{x}_i^{\mathrm{adv }} ; \theta\right), y_i\bigr)- \ell\bigl(f\left(\boldsymbol{x}_i ; \theta\right), y_i\bigr)|}{\ell\bigl(f\left(\boldsymbol{x}_i ; \theta\right), y_i\bigr) + \delta},
\label{eq:adm}
\end{align}
where $\delta$ is a small constant for numerical stability. Intuitively, this measure considers a sample to be more difficult if a small adversarial perturbation causes a large increase in loss. This makes it possible to identify relatively vulnerable samples, even in regions where traditional difficulty metrics become ineffective due to flatness.

\subsection{Curriculum Update}
Based on the estimated difficulty defined in Eq.~\eqref{eq:adm}, we update the curriculum in the same manner as in SPL. The sample weight $v_i^{(t+1)}$ at epoch $t+1$ is defined as:
\begin{align}
v_i^{(t+1)}= \begin{cases}1, & \xi_i \leq \lambda^{(t)} \\ 0, & \text { otherwise }\end{cases}.
\end{align}
Here, $\lambda^{(t)}$ is a threshold parameter at epoch $t$. By gradually increasing $\lambda^{(t)}$ over the course of training, the model begins with samples that have low ADM scores (i.e., easier examples) and progressively incorporates more difficult ones. Since ADM serves as a difficulty estimator, it easily adapt to symmetric curriculum schedules~\cite{sspl}, proposed in SSPL.

In this way, ADM bridges the gap between CL and flatness-aware optimization by enabling reliable sample selection even as the model converges toward flatter regions. As a result, our method effectively combines the strengths of CL in progressive training and SAM in guiding the optimization toward flat minima.

\section{Evaluation}
\label{sec:eval}
To validate the effectiveness of our proposed method, we design an experimental protocol that systematically compares three key components: the use of CL, the use of SAM, and the integration of both.
Specifically, for CL, we evaluate ERM as baseline and curriculum-based methods including SPL~\cite{spl} and SSPL~\cite{sspl}. CL-based methods are implemented using the weighted loss (Eq.~\eqref{eq:loss_spl}), and the difficulty score is determined either by the loss, the input gradient norm  (proposed in SSPL as the Gradient Difficulty Measure, or GDM), or our proposed ADM. Each of these variants is optimized using SGD and SAM~\cite{sam}.
This experimental design enables a comprehensive evaluation of how curriculum-based sample selection and flatness-aware optimization individually and jointly contribute to generalization performance. All results are reported as the mean and standard deviation of test accuracy over three models trained with different random seeds.

\subsection{Image Classification}
\label{sec:imgcls}
\noindent\textbf{Experimental Setup:}
We evaluate our method on image classification using the \texttt{CIFAR10/100} datasets~\cite{cifar}. These datasets are widely used in both the CL and flatness-aware optimization settings. For training, we follow the hyperparameter settings reported in prior work~\cite{sam}. The learning rate is initially set to 0.1 and decayed by a factor of 0.2 at epochs 60, 120, and 160. We use a batch size of 128 and train for 200 epochs. As model architectures, we adopt ResNet18~\cite{resnet}, WideResNet28-2 (WRN28\_2), and WideResNet28-10 (WRN28\_10)~\cite{wrn}, all of which are trained from scratch. For data augmentation, we apply the standard scheme: random cropping and random horizontal flipping.

\acccifar

\noindent\textbf{Experimental Result:}
Table~\ref{tb:acccifar} shows the accuracies on \texttt{CIFAR10/100}. Overall, the CL strategy based on the proposed ADM achieved performance that is comparable to or better than CL based on conventional difficulty measures such as loss or input gradient norm (GDM). In particular, when combined with SAM, ADM yields the highest accuracy across various model configurations including ResNet and WideResNet, indicating strong compatibility with SAM. For instance, under the SSPL+ADM+SAM setting on \texttt{CIFAR100}, ADM achieves $79.89_{\pm 0.27}$ with ResNet18, $77.05_{\pm 0.15}$ with WRN28-2, and $83.24_{\pm 0.04}$ with WRN28-10, outperforming all other difficulty measures in each case. On \texttt{CIFAR10}, ADM demonstrates comparable or superior generalization across all configurations and tends to achieve the best accuracy when combined with SAM, while maintaining a similar performance to loss-based and GDM-based curricula.

These results indicate that ADM functions as a stable and effective difficulty measure for image classification tasks, offering competitive generalization without performance degradation, especially when used alongside SAM.

\subsection{Corruption Robustness}
\noindent\textbf{Experimental Setup:}
To further analyze the effectiveness of our method, we evaluate model corruption robustness using \texttt{CIFAR10-C} and \texttt{CIFAR100-C}~\cite{cifarc}, which contain 15 types of common corruptions including noise, blur, weather, and digital artifacts. Based on this experiment, we confirm that models converging to flatter minima exhibit higher robustness to image corruption. We use the models trained in Section~\ref{sec:imgcls} for inference.

\cifarc

\noindent\textbf{Experimental Result:}
The results are summarized in the Table~\ref{tb:cifar_c}. Overall, models trained with SAM consistently achieve higher robustness across all curriculum strategies compared to their SGD-based counterparts. In particular, SSPL combined with the proposed ADM consistently yields the highest or comparable performance across both ResNet and WideResNet architectures. This result highlights the strong compatibility between SAM and ADM-based CL. Notably, SSPL+ADM with SAM achieves $76.00_{\pm 0.26}$ on \texttt{CIFAR10-C} and $50.95_{\pm 0.37}$ on \texttt{CIFAR100-C} (ResNet-18 backborne), confirming the ability of our method to maintain robustness under common image corruptions.

\subsection{Fine-grained Image Classification}
\noindent\textbf{Experimental Setup:}
Next, we discuss the effectiveness of our method from the perspective of CL on fine-grained image classification. We conduct comparative experiments on four benchmarks: Stanford Cars (\texttt{Cars})~\cite{cars}, FGVC-Aircraft (\texttt{Aircraft})~\cite{aircraft}, CUB-200-2011 (\texttt{CUB})~\cite{cub}, and Food-101 (\texttt{Food})~\cite{food}. Following the protocol used in the standard classification experiments, we adopt ResNet18~\cite{resnet}. To clearly evaluate the effect of CL, all models are trained from scratch without pretraining. For data augmentation, we follow the standard ImageNet preprocessing~\cite{imagenet}. For each dataset, we select the batch size from $\{32, 64, 128, 256\}$ that yielded the best performance. Specifically, we used a batch size of 128 for \texttt{Food}, and 32 for the other datasets. The learning rate, scheduler, and other hyperparameters follow the same settings as in the standard image classification experiments. For the radius of SAM, we set $\rho = 0.05$.

\finegrainedacc

\noindent\textbf{Experimental Result:}
Table~\ref{tb:fine_grained_acc} shows the classification results for each fine-grained dataset.  Fine-grained recognition requires distinguishing subtle differences between visually similar classes. In such tasks, curriculum learning is particularly effective, as it enables the model to first capture simple and common patterns in early training stages and then focus on harder, more discriminative features later. From the table, ADM supports this strategy by providing a stable and flatness-consistent difficulty measure.

The benefits of ADM are especially evident when combined with SSPL under SAM optimization. For example, SSPL+ADM achieves $83.85_{\pm 0.05}$ on \texttt{Cars}, $63.44_{\pm 0.24}$ on \texttt{CUB}, and $82.01_{\pm 0.09}$ on \texttt{Food}, outperforming other difficulty measures on these datasets. These results demonstrate that ADM is well-suited for fine-grained image classification and can effectively enhance generalization when used in conjunction with flat-minima optimization methods like SAM.

\subsection{Domain Generalization}
\noindent\textbf{Experimental Setup:}
In this section, we examine the domain generalization capability of our method from the perspective of flat minima. Previous studies~\cite{sagm,swad} have shown that flat minima tends to generalize better to unseen domains. Our goal is to verify whether this property is preserved in our method, which integrates CL with SAM. To this end, we follow the standard protocol~\cite{dgeval,rethinkingdg} and evaluate our method on four widely used datasets: \texttt{PACS}~\cite{pacs}, \texttt{VLCS}~\cite{vlcs}, \texttt{Office-Home} (\texttt{OH})~\cite{officehome}, and \texttt{TerraIncognita} (\texttt{TI})~\cite{ti}. For feature extraction, we adopt a ResNet50 backbone pretrained on ImageNet~\cite{imagenet} using MoCo-v2~\cite{mocov2} in a self-supervised manner. Evaluation is performed using the leave-one-domain-out cross-validation setting~\cite{dgeval}, where one domain is held out as the target, and the remaining domains are used as training sources.
We set the SAM perturbation radius $\rho$ to 0.05, following the original SAM configuration. Other hyperparameters are configured according to SAGM~\cite{sagm}.

\dgacc

\noindent\textbf{Experimental Result:}
Table~\ref{tb:dg} shows the classification accuracy results for each domain generalization dataset. As shown in the table, even when using SAM, applying our proposed ADM within SSPL achieves better average accuracy across all four datasets, comparable to or surpassing that of methods without SAM, such as SSPL with Adam. In particular, on \texttt{TerraIncognita} (\texttt{TI}), our method exhibits a notable improvement over other baselines.

The effectiveness of SAM has been previously reported in domain generalization settings. Our results further confirm that this benefit persists even when combined with CL. The proposed ADM contributes to this by facilitating a sample selection strategy that emphasizes stable learning early on and progressively shifts toward more challenging cases, which is particularly effective in the presence of distribution shifts.

\subsection{Visualization of Difficulties}
To assess the effectiveness of our criterion, we visualize the per-sample difficulty used for curriculum learning during training on \texttt{CIFAR10} with SGD and SAM (Figure~\ref{fig:sharp_flat_difficulty}). As optimization proceeds, both loss- and gradient-based measures shrink and become nearly uniform near flat minima, making curricula based on them increasingly uninformative. Notably, with conventional difficulties this phenomenon is especially evident when optimizing with SAM, but it also appears with SGD, highlighting the limitations of these measures. In contrast, our adversarial vulnerability-based measure (ADM) maintains separability among samples and continues to provide a reliable signal for curriculum decisions even in the late stage of training.

\subsection{Flatness Analysis}
\acchessian
Finally, we investigate the relationship between our ADM-based CL and flatness. To this end, we analyze the ResNet18 model w/ ADM-based CL trained in Section~\ref{sec:imgcls}, using both quantitative and qualitative approaches. Specifically, we measure the maximum eigenvalue of the Hessian matrix~\cite{hessian} and visualize the loss landscape~\cite{losslandscape}. Table~\ref{tb:acc_hessian} reports the maximum Hessian eigenvalues, while Figure~\ref{fig:loss_landscape} presents the loss landscape.

As shown in the table, applying ADM-based CL with SAM results in consistently smaller eigenvalues compared to standard ERM and SPL across both datasets. On \texttt{CIFAR100}, for example, the eigenvalue for SAM with ADM-based SPL is reduced to 7.49, compared to 9.47 in ERM. Similarly, on \texttt{CIFAR10}, the combination of ADM and SAM yields the lowest value of 2.35 among all settings. These reductions suggest that, even without directly minimizing sharpness, our ADM-based CL implicitly promotes convergence toward flatter regions of the loss surface as a byproduct of its sample selection strategy. On the other hand, the results also suggest that SSPL scheduling strategies may be associated with sharper minima. This observation highlights the importance of designing curriculum schedules that are not only effective for progressive sample selection but also compatible with flatness-aware optimization.

\visdiff
\losslandscape

\section{Conclusion}
In this work, we explored the potential of CL during the optimization process toward flatter minima. We identified a key limitation in conventional difficulty measures, which become ineffective when both loss values and gradient norms decrease uniformly in flat regions. To address this issue, we proposed the Adversarial Difficulty Measure (ADM), which evaluates sample difficulty based on adversarial vulnerability. Our method enables reliable curriculum scheduling even as the model converges to flatter region. This allows CL and SAM to be effectively integrated. The experimental results demonstrated that ADM-based CL outperforms conventional CL while preserving flatness, as supported by Hessian eigenvalue analysis and loss landscape visualizations.

\bibliography{main}
\end{document}